\documentclass[10pt,twocolumn,letterpaper]{article}

\usepackage{wacv}              %
\usepackage{cuted}
\usepackage{algorithm}
\usepackage{multirow}
\usepackage{fancyvrb}

\usepackage{algpseudocode}
\algrenewcommand\algorithmicrequire{\textbf{Input:}}
\algrenewcommand\algorithmicensure{\textbf{Output:}}

\definecolor{wacvblue}{rgb}{0.21,0.49,0.74}
\usepackage[pagebackref,breaklinks,colorlinks,allcolors=wacvblue]{hyperref}

\def\wacvPaperID{1225} %
\def\confName{WACV}
\def\confYear{2027}

\title{InteractAgent: Agentic Human Motion Interaction with Memory-Augmented Reflection}

\author{Senhe Zhang$^{1*}$\quad
Tao Zhang$^{1*}$\quad
Yujia Zhang$^{1}$\quad
Zeyu Zhang$^{2\dag}$\quad
Dong Gong$^{2\ddag}$\\
$^1$AI Geeks\quad
$^2$UNSW Sydney\\
\small $^*$Equal contribution.
$^\dag$Project lead.
$^\ddag$Corresponding author: dong.gong@unsw.edu.au.}

\begin{document}
\maketitle

\begin{strip}
\centering
\includegraphics[width=\textwidth]{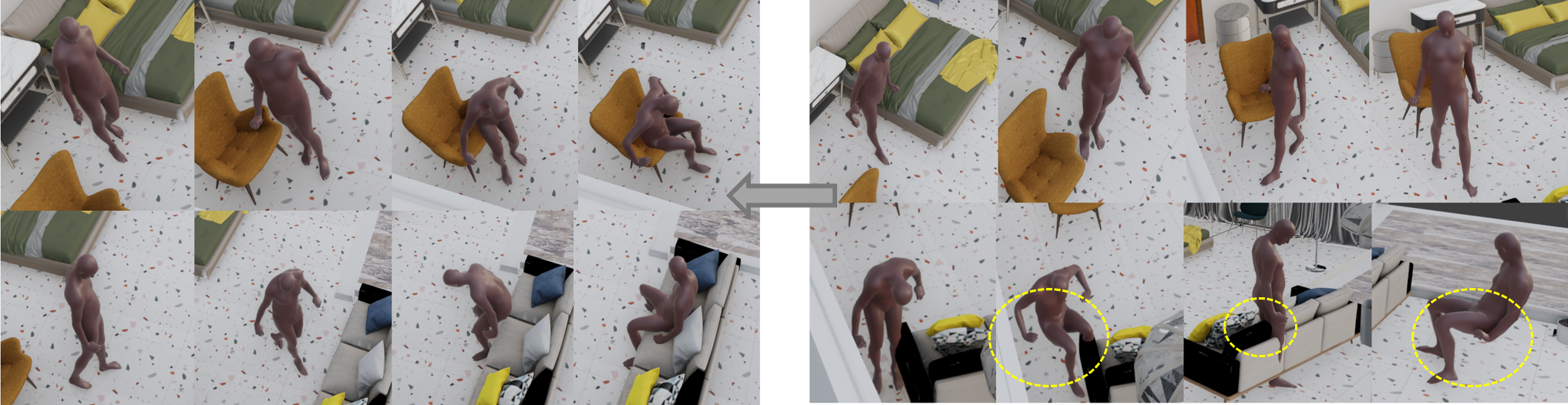}
\captionof{figure}{
\textbf{InteractAgent vs.\ Naive:}
\textit{InteractAgent} (left) couples \textbf{Online Observation} with \textbf{Memory-Augmented Reflection},
iteratively refining motion plans to achieve long-horizon, physically plausible
human--scene interactions from images and language only.
A naive text-to-motion baseline (right) lacks feedback and quickly accumulates errors,
resulting in severe penetration and incomplete tasks.
}
\label{fig:teaser}

\end{strip}

\begin{abstract}Long-horizon human-scene interaction requires coordinated spatial reasoning, physical feasibility, and the ability to recover from compounding execution errors.
Text-driven motion models are largely scene-agnostic and lack closed-loop correction, while geometry-conditioned approaches depend on paired scene--motion training or explicit 3D preprocessing that limits generalization to novel layouts and open-vocabulary goals.
We present \textbf{InteractAgent}, an agentic multimodal LLM system for long-horizon human-scene interaction that \textbf{plans, executes, observes, and reflects} in a closed loop, using a single MLLM as both planner and evaluator and a frozen MotionLLM as the executor.
InteractAgent integrates \textbf{Online Observation} for step-wise recovery decisions (\textit{continue/redo/replan}) and \textbf{Memory-Augmented Reflection} for multi-round plan refinement with adaptive stopping.
On UniHSI, InteractAgent improves task success and reduces penetration compared with training-free baselines.
Code: \url{https://github.com/AIGeeksGroup/InteractAgent}.  
\end{abstract}

\section{Introduction}

Understanding and synthesizing human behavior within complex environments is fundamental to embodied artificial intelligence. A key challenge lies in generating \textit{long-horizon human--scene interactions}: temporally extended motion sequences where agents navigate, manipulate objects, and accomplish tasks in structured 3D spaces. Such scenarios require joint reasoning over spatial layouts, physical constraints, and semantic goals while maintaining temporal coherence. Recent advances in human motion generation have achieved impressive progress through large-scale pretraining~\cite{guo2022generating,jiang2023motiongpt,zhang2023t2m}. However, these models typically operate in \textit{scene-agnostic} latent spaces---motion is generated independently of environmental context, lacking awareness of object locations, affordances, or contact constraints. A generated ``sit on the chair'' motion may exhibit plausible kinematics in isolation, yet produce severe penetration when placed in a real scene. 

To incorporate scene awareness, prior work has introduced explicit geometric constraints and geometry-conditioned models~\cite{hassan2021populating,xiao2023unified,jiang2024scaling}. These approaches can provide accurate scene grounding, but may require paired scene--motion annotations or specialized training. We instead investigate training-free planning and feedback around pretrained models. InteractAgent couples a pretrained multimodal large language model (MLLM) with a frozen MotionLLM executor to generate scene-aware motions from visual observations and natural-language instructions. The MLLM produces action plans, observes rendered execution outcomes, and revises its plans without task-specific finetuning. The intermediate plans and feedback are human-readable; we do not claim that they faithfully explain the internal reasoning of the underlying models.

The central insight is that \textit{reflection}---observing outcomes, diagnosing errors, and iteratively refining plans---provides a powerful inductive bias for long-horizon interaction. We introduce two complementary mechanisms. \textbf{Online Observation} performs step-wise feedback: after each action, the MLLM evaluates the rendered motion and decides whether to continue, redo, or replan, preventing early errors from propagating. \textbf{Memory-Augmented Reflection} operates at the task level by accumulating planning--execution pairs across attempts and refining strategies through multi-round reasoning with an adaptive stopping criterion. Together, these mechanisms bridge \textit{local robustness} and \textit{global optimization}, enabling self-improvement without supervision.

These observations motivate the following question: can structured visual feedback and cross-attempt reflection improve long-horizon human--scene interaction without task-specific training? We study this question by treating the pretrained MLLM and motion executor as fixed components and organizing their use through step-wise recovery and trajectory-level refinement. The planner operates on rendered visual observations rather than explicit geometric features, while the experimental pipeline still uses 3D scenes for rendering and geometric penetration evaluation.

\paragraph{Contributions.}
In this work we propose \textbf{InteractAgent}, a training-free reflective framework for generating long-horizon human--scene interactions.
Our main contributions are:
\begin{itemize}[leftmargin=*,noitemsep,topsep=0pt]
    \item \textbf{Training-free perception-to-action pipeline.}
    We integrate scene understanding, task decomposition, and motion synthesis solely through multimodal prompting---requiring no geometry-conditioned training or task-specific finetuning. A single MLLM serves as the unified planner and evaluator.
    \item \textbf{Dual-layer reflection for robust long-horizon planning.}
    We introduce \emph{Online Observation} for step-wise closed-loop correction and \emph{Memory-Augmented Reflection} for multi-round task-level optimization, jointly stabilizing long-horizon execution.
    \item \textbf{Adaptive stopping for efficient refinement.}
    An adaptive stopping mechanism automatically determines the number of reflection rounds, improving efficiency while preserving global consistency.
    \item \textbf{Extensive evaluation on UniHSI and TRUMANS.}
    We show that InteractAgent achieves competitive success rates and significantly lower penetration errors than training-free and single-pass baselines, while exposing human-readable intermediate plans and feedback for inspection.
\end{itemize}

\section{Related Work}

\paragraph{Human motion generation.}
Human motion generation aims to synthesize realistic, temporally coherent, and semantically meaningful motion under text, audio, or control signals.
Early work relied on recurrent or transformer-based models trained on MoCap datasets.
Text-to-motion methods such as Guo et al.~\cite{guo2022generating} and T2M-GPT~\cite{zhang2023t2m} learn a shared latent space or VQ-VAE codebook to map text to discrete motion tokens, while MotionGPT~\cite{jiang2023motiongpt} unifies motion and text as a single “language”.
MoMask~\cite{guo2024momask} improves temporal coherence via hierarchical quantization and masked motion modeling.
Diffusion-based approaches further enhance realism and controllability: Pi et al.~\cite{pi2023hierarchical} use hierarchical diffusion for detailed human--object motion, and Wang et al.~\cite{wang2025cedt2m} propose a cross-modal encoder--decoder mixture for text-driven 3D motion.
However, these models operate in scene-agnostic latent spaces and do not explicitly account for environmental constraints, contacts, or navigation, limiting their use in embodied, scene-grounded settings.

\paragraph{Human motion interaction.}
To go beyond single-agent motion, prior work studies human--human (HHI), human--object (HOI), and human--scene (HSI) interactions.
HOI frameworks~\cite{li2023object,taheri2020grab,li2024controllable} model affordances and contact dynamics through object-conditioned synthesis, grasping datasets, or controllable interaction goals.
HSI methods~\cite{hassan2021populating} incorporate 3D layouts and pose--scene embeddings to generate physically plausible motion in structured environments.
A variety of datasets and simulators support grounded interaction learning:
BEHAVE~\cite{bhatnagar2022behave} captures multi-view RGB-D human--object interactions, HOISim~\cite{zakour2021hoisim} provides simulation-based HOI data, and Wang et al.~\cite{wang2021synthesizing} generate long-term motion-in-scene sequences.
Holistic++~\cite{chen2019holistic++} jointly infers layout, human pose, and affordances, while Wu et al.~\cite{wu2025human} map natural-language instructions to 3D interactive motions.
These approaches significantly improve realism but rely on hand-crafted scene features or geometry-conditioned training, and they lack the high-level semantic reasoning needed for adaptive planning in novel environments.

\paragraph{Vision-language models for scene understanding.}
Vision-language models (VLMs) and multimodal LLMs (MLLMs) enable cross-modal reasoning and grounding between perception and language.
BLIP-2~\cite{li2023blip} and PaLM-E~\cite{driess2023palm} align visual encoders with large language models for semantic localization, reasoning, and action grounding.
VIMA~\cite{jiang2022vima} and RT-2~\cite{zitkovich2023rt} further demonstrate that multimodal models can serve as general-purpose planners for embodied control.
Qwen-VL~\cite{bai2023qwen} improves high-resolution visual reasoning and spatial understanding.
3D-aware extensions such as Scene-LLM~\cite{fu2024scene} and 3D-LLM~\cite{hong20233d} inject 3D geometry for 3D-aware reasoning and manipulation.
Chen et al.~\cite{chen2025vision} use VLMs to mine pose knowledge for human pose estimation, while EgoNav~\cite{wang2024egonav}, UniEgoMotion~\cite{patel2025uniegomotion}, and Wang et al.~\cite{wang2023scene} tackle egocentric motion forecasting and pose estimation by leveraging human--scene relationships.
Human-centric metrics~\cite{fu2020human} quantify spatial affordance and reachability.
Together, these works suggest that coupling a grounding-capable MLLM with tokenized motion models (e.g., MotionGPT, T2M-GPT, MoMask) can support a perception--planning--generation pipeline.
Our framework builds on this idea and explicitly connects MLLM-based scene analysis, reflective planning, and a MotionLLM executor to achieve end-to-end motion generation from visual perception to 3D realization.

\section{Method}
\label{sec:method}

In InteractAgent, we rely on a \textbf{single multimodal large language model (MLLM)} that accepts both images and text and produces structured outputs in natural language.
For clarity of exposition, we describe separate ``VLM'' and ``LLM'' roles (for scene analysis, plan generation, evaluation, and reflection), but in practice \emph{all} these roles are instantiated by the same underlying MLLM backbone.
Unless otherwise specified, we use GPT-4o as this backbone in all experiments and study alternative backbones (e.g., Qwen-VL-Max) only in ablations.

For motion synthesis, we use a separate \textbf{MotionLLM executor} that maps a sequence of natural-language actions to a full 3D human motion trajectory.
In all our experiments, this executor $Q$ is instantiated with the MotionLLM module from the Motion-Agent framework, which combines a Gemma-2B language backbone with a VQ-VAE motion decoder and SMPL-X reconstruction.
We treat $Q$ as a fixed component: no finetuning or modification is performed on this module, and our contributions focus on how to \emph{plan} and \emph{reflect} around it.

\begin{figure*}[t]
    \centering
    \includegraphics[width=0.9\textwidth]{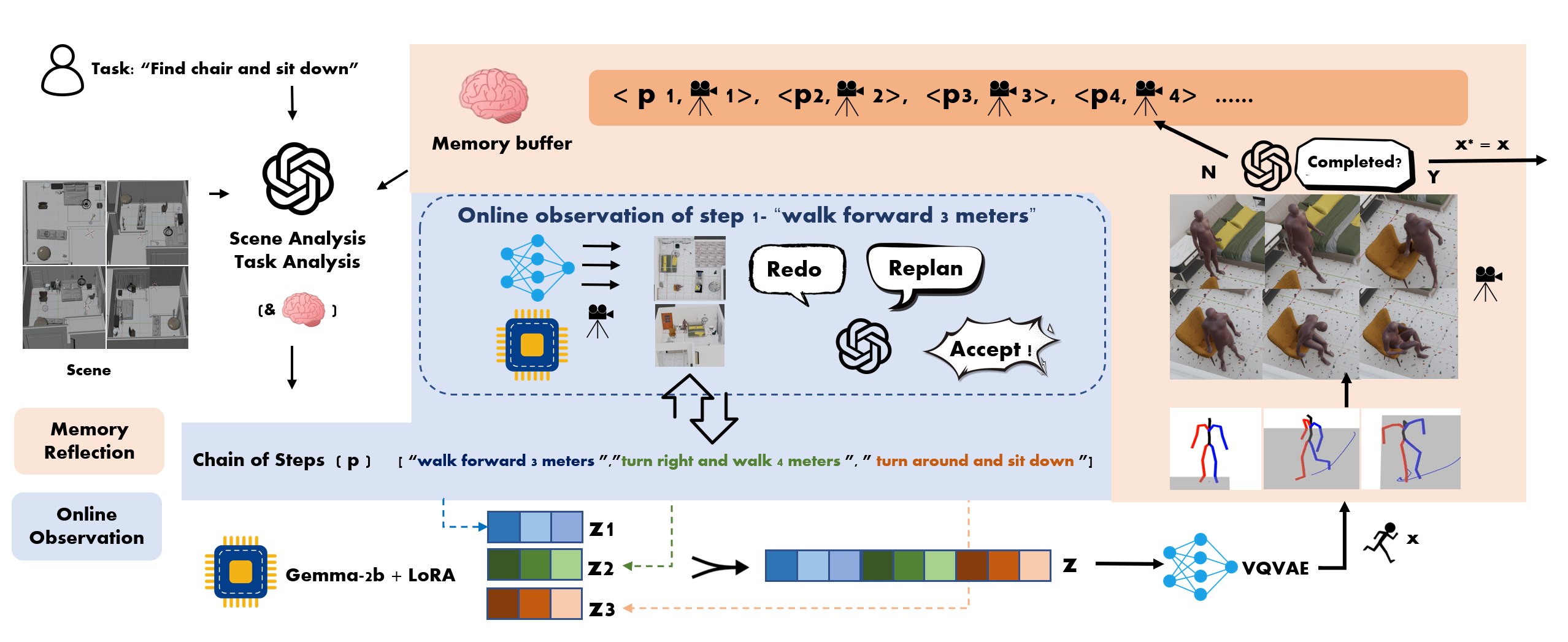}
    \caption{\textbf{Overview of InteractAgent.} 
Our framework couples a single multimodal LLM with a frozen MotionLLM executor. 
Two key components enable training-free, long-horizon interaction generation: 
(1) \emph{Online Observation} provides step-wise visual feedback during execution, 
allowing the MLLM to correct local deviations; 
(2) \emph{Memory-Augmented Reflection} aggregates planning--execution histories across 
rounds and refines future plans at the trajectory level. 
Together, these mechanisms form a dual-layer reflective loop that progressively 
improves motion quality and scene consistency.}
    \label{fig:trajectory}
\end{figure*}

\subsection{Overview}

InteractAgent bridges visual perception and human motion synthesis by
explicitly grounding motion planning in the surrounding scene. Given one or
more scene images and a high-level task description, the system first performs
MLLM-based scene analysis, then decomposes the task into atomic actions, and
finally invokes a frozen MotionLLM executor to synthesize a full 3D motion
sequence.

Concretely, a single multimodal LLM handles four roles: scene understanding,
action sequence generation, step-wise online observation, and memory-augmented
reflection. The MotionLLM is used purely as an executor that converts
natural-language action descriptions into SMPL-X motion trajectories. Our
contributions therefore focus on how to plan and reflect around this fixed
executor via dual-layer reflection, rather than designing new motion
architectures.

\begin{figure}[t]
  \centering
  \includegraphics[width=\columnwidth]{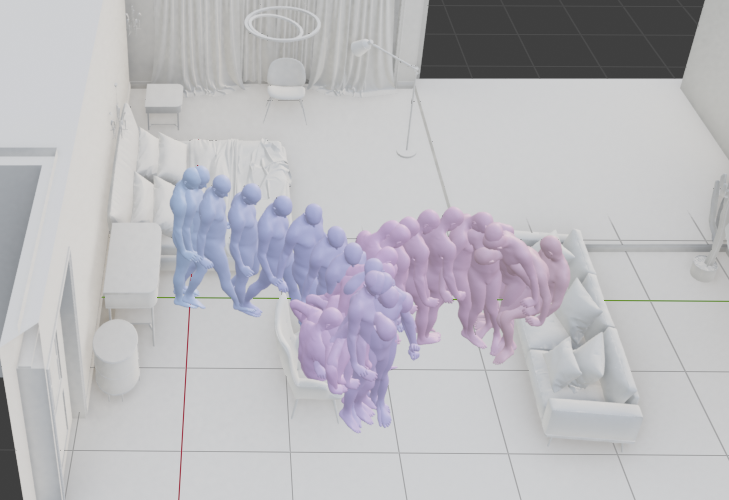}
  \caption{Trajectory visualization image used as a compact reflection signal for the MLLM.}
  \label{fig:tra}
\end{figure}

\subsection{Scene Motion Planning Pipeline}

InteractAgent establishes an end-to-end pipeline that connects visual perception with motion synthesis.
Given a scene observation $\mathbf{s}$ (e.g., RGB images or multi-view renderings) and a high-level task description $\mathbf{t}$, the planner produces a structured motion sequence $\mathbf{x} = \{\mathbf{x}_1,\ldots,\mathbf{x}_L\}$ that realizes the intended behavior within the environment.

We encode the long-horizon interaction history as a single trajectory visualization image (Fig.~\ref{fig:tra}) and use it as a compact reflection signal for the MLLM.

Formally, we denote the underlying motion executor by $Q$, which maps
a sequence of natural-language action descriptions to a full motion
trajectory:
\[
Q:\; \mathbf{p} = \{\mathbf{p}_1, \ldots, \mathbf{p}_K\} \;\longrightarrow\; \mathbf{x}.
\]
For clarity, we omit the implicit dependence of $Q$ on the scene
$\mathbf{s}$ and task $\mathbf{t}$, and refer to it simply as the
MotionLLM executor used throughout Algorithms~\ref{alg:online_observation}
and~\ref{alg:memory_reflection}.

\paragraph{Scene Understanding.}
To initialize the process, the scene is first provided as a prompt to the MLLM, which interprets spatial layouts, object affordances, and potential interaction sites:
\[
\mathbf{g} = f_{\mathrm{MLLM}}(\mathbf{s}).
\]
The scene representation $\mathbf{g}$ encodes entities, relations, and spatial structure for downstream reasoning.

\paragraph{Action Sequence Generation.}
Conditioned on $\mathbf{g}$ and the task prompt $\mathbf{t}$, the same MLLM decomposes the task into a sequence of atomic, executable natural-language actions:
\[
\mathbf{p} = \{\mathbf{p}_1,\ldots,\mathbf{p}_K\},\quad \mathbf{p}_k \in \text{NL}.
\]
Each action describes a short, unambiguous motion primitive (e.g., ``turn left'', ``walk forward 1.5 meters'').

\paragraph{Action-to-Motion Execution.}
We do not design a new motion synthesis model.
Instead, InteractAgent directly calls the frozen MotionLLM executor to realize the entire action list:
\[
\mathbf{x} = Q(\mathbf{p}).
\]
Internally, MotionLLM encodes the natural-language sequence, predicts motion tokens, decodes them with a VQ-VAE motion decoder, and reconstructs SMPL-X parameters.
We treat this process as an internal implementation detail of $Q$ and focus on how to generate and refine $\mathbf{p}$ via reflection.

This pipeline constitutes the forward execution path of InteractAgent---from perceptual grounding to motion realization on top of a fixed MotionLLM executor.
To enhance robustness through step-wise feedback, we introduce Online Observation (Section~\ref{subsec:online_obs}) and Memory-Augmented Reflection (Section~\ref{subsec:memory_reflection}), both implemented via the same MLLM backbone.

\subsection{Online Observation}
\label{subsec:online_obs}

While the scene motion planning pipeline provides a structured perception-to-action flow, it executes the entire action sequence in an open-loop manner, assuming all predicted actions are correct once generated.
In practice, early-step inaccuracies (e.g., over- or under-estimated displacements) often propagate through subsequent actions, resulting in physically inconsistent or semantically implausible motions.
Such forward error accumulation not only degrades overall realism but also complicates global reflection, as it becomes difficult to isolate whether the degradation stems from local planning errors or model stochasticity.

To alleviate this issue, we introduce an \textbf{Online Observation} mechanism that integrates step-wise reflection and local replanning during execution.
After each atomic action $\mathbf{p}_k$ is generated and executed by the MotionLLM executor,
\[
\mathbf{x}_k = Q(\mathbf{p}_k),
\]
the resulting visual outcome $\mathbf{v}_k$ is rendered (e.g., as a short clip or keyframes) and evaluated through the same MLLM, forming an observation pair $(\mathbf{p}_k, \mathbf{v}_k)$.

The MLLM evaluates the alignment between planned action $\mathbf{p}_k$ and rendered outcome $\mathbf{v}_k$, producing a score and textual feedback:
\[
(\mathrm{score}_k, \mathrm{feedback}_k) = f_{\mathrm{MLLM}}(\mathbf{p}_k, \mathbf{v}_k).
\]

Based on this feedback, the language interface of the MLLM makes a discrete decision
\[
d_k \in \{\text{continue}, \text{redo}, \text{replan}\},
\]
corresponding to one of three strategies:
\begin{itemize}
    \item \textbf{Continue:} if $\mathrm{score}_k > \tau_{\text{accept}}$, accept the current action $\mathbf{p}_k$ and proceed with the next planned step;
    \item \textbf{Redo:} detect misalignment in the current action, locally correct and regenerate $\mathbf{p}_k$ before continuing the sequence;
    \item \textbf{Replan:} detect trajectory drift, accept $\mathbf{p}_k$ but regenerate the remaining actions
    $\{\mathbf{p}_{k+1}, \ldots, \mathbf{p}_K\}$ conditioned on the updated scene state.
\end{itemize}

Through this process, each step produces an optimized action $\mathbf{p}^\ast_k$, and the planner maintains an updated list of refined actions
\[
\mathbf{p}^\ast = \{\mathbf{p}^\ast_1, \mathbf{p}^\ast_2, \ldots, \mathbf{p}^\ast_K\}.
\]
This list captures the locally adjusted and globally consistent plan emerging from the online reflection loop.
After completing all steps, the optimized sequence is executed once more by the MotionLLM executor to obtain the final trajectory:
\[
\mathbf{x}^\ast = Q(\mathbf{p}^\ast).
\]

The full procedure is summarized in Algorithm~\ref{alg:online_observation}.

\begin{algorithm}[t]
\caption{Online Observation (Step-wise)}
\label{alg:online_observation}
\begin{algorithmic}[1]
\Require Scene $\mathbf{s}$; initial task plan $\mathbf{p} = \{\mathbf{p}_1,\ldots,\mathbf{p}_K\}$;
         MotionLLM executor $Q$; MLLM evaluator $f_{\mathrm{MLLM}}$;
         max retries per step $R_{\max}$.
\Ensure Refined action list $\mathbf{p}^\ast$, final motion $\mathbf{x}^\ast$

\State $\mathbf{p}^\ast \leftarrow [\,]$
\For{$k = 1$ to $K$}
    \State $\text{retry\_count} \gets 0$  \Comment{Initialize retry counter}
    \Repeat
        \State $\mathbf{x}_k \leftarrow Q(\mathbf{p}_k)$
        \State $\mathbf{v}_k \leftarrow \mathrm{Render}(\mathbf{x}_k)$
        \State $(\mathrm{score}_k, \mathrm{feedback}_k) \leftarrow f_{\mathrm{MLLM}}(\mathbf{p}_k,\mathbf{v}_k)$
        \State $d_k \in \{\text{continue},\text{redo},\text{replan}\}$ based on $\mathrm{feedback}_k$
        \If{$d_k \neq \text{continue}$}
            \State $\text{retry\_count} \gets \text{retry\_count} + 1$
        \EndIf
        \If{$d_k = \text{redo}$}
            \State regenerate $\mathbf{p}_k$ using MLLM feedback
        \ElsIf{$d_k = \text{replan}$}
            \State replan $\{\mathbf{p}_{k+1}, \ldots, \mathbf{p}_K\}$ conditioned on current state
        \EndIf
    \Until{$d_k = \text{continue}$ \textbf{or} $\text{retry\_count} \ge R_{\max}$}
    \State $\mathbf{p}^\ast_k \leftarrow \mathbf{p}_k$  \Comment{Accept last attempt}
\EndFor
\State $\mathbf{x}^\ast \leftarrow Q(\mathbf{p}^\ast)$
\State \Return $\mathbf{p}^\ast,\,\mathbf{x}^\ast$
\end{algorithmic}
\end{algorithm}

By embedding reflection within each execution step, InteractAgent continuously aligns its planned actions with the observed outcomes, forming a closed-loop perception--action cycle.
This design transforms the planner from an offline generator into an \emph{execution-aware interactive system} that perceives, acts, and self-adjusts in real time.
To prevent excessive local oscillation, Online Observation enforces a maximum number of correction attempts for each action step.
If repeated \textit{redo} or \textit{replan} decisions exceed this limit, the system terminates early and accepts the most recent action $\mathbf{p}_k$ as the final outcome for that step.
Once the limit is reached, the planner does not attempt additional adjustments and proceeds using only the last generated version of the action, discarding previous failed attempts.

While Online Observation provides stable step-wise correction, it remains limited to short-horizon improvements.
To further enhance long-term consistency and task-level reasoning, we introduce a \textbf{Memory-Augmented Reflection} mechanism.

\subsection{Memory-Augmented Reflection}
\label{subsec:memory_reflection}

Online Observation focuses on local corrections within a single execution of a task.
However, long-horizon behaviors often require global adjustments that can only be discovered by comparing \emph{multiple} executions of the same task.
To this end, we augment InteractAgent with a trajectory-level memory that accumulates planning--execution pairs across attempts and enables global reflection.

\paragraph{Iterative Reflection Process.}
Given a specific task defined by scene $\mathbf{s}$ and goal $\mathbf{t}$, the model executes the complete planning and motion generation process multiple times.
Each execution produces a planned action sequence $\mathbf{p}_i$ and the corresponding rendered outcome $\mathbf{v}_i$, which form a planning--result pair $(\mathbf{p}_i, \mathbf{v}_i)$.
These pairs are accumulated as memory:
\[
\mathcal{M}_i = \{(\mathbf{p}_1, \mathbf{v}_1), (\mathbf{p}_2, \mathbf{v}_2), \ldots, (\mathbf{p}_i, \mathbf{v}_i)\}.
\]
This memory encapsulates both successful and suboptimal past attempts, providing visual evidence of how different planning strategies manifest as physical motion.

At the $(i+1)$-th iteration, the model performs reflection by reasoning over the accumulated memory $\mathcal{M}_i$ when generating a new plan:
\[
\mathbf{p}_{i+1} = f_{\mathrm{reflect}}(\mathcal{M}_i, \mathbf{s}, \mathbf{t}),
\]
where $f_{\mathrm{reflect}}$ is implemented by prompting the same MLLM $f_{\mathrm{MLLM}}$ with the task description, scene context, and all previous planning--execution pairs.
The MLLM can directly observe rendered videos $\{\mathbf{v}_1, \ldots, \mathbf{v}_i\}$ to identify patterns of failure---such as persistent penetration errors, trajectory drift, or incomplete task execution---and adjust its planning strategy accordingly.

The newly generated plan $\mathbf{p}_{i+1}$ is then executed and rendered as $\mathbf{v}_{i+1}$, after which the pair is appended to memory:
\[
\mathcal{M}_{i+1} = \mathcal{M}_i \cup \{(\mathbf{p}_{i+1}, \mathbf{v}_{i+1})\}.
\]
Through this iterative process, the memory gradually accumulates planning--execution correspondences that guide the model toward more coherent and physically consistent motion generation.

Importantly, unlike step-wise Online Observation---which only retains the most recent attempt when its retry limit is reached---the memory module preserves \emph{all} planning--execution pairs collected across iterations.
Even if the system reaches the maximum number of reflection rounds and the task remains incomplete, the full memory $\mathcal{M}_i$ is kept intact.
This enables the reflection model to exploit the entire trajectory of past successes, failures, and partial improvements, rather than relying solely on the final attempt.

\paragraph{Adaptive Stopping Criterion.}
After each reflection round, the evaluator checks the rendered execution against the task requirements. A \textit{Pass} judgment terminates refinement; otherwise, the system continues until the reflection budget is exhausted. We use $N_{\max}=5$ as the default budget.

Early termination avoids executing unused reflection rounds once the evaluator judges the task complete. However, this visual judgment is not a guarantee of geometric or physical correctness. We therefore report body--scene penetration separately from task success. The selected budget is an experimental setting rather than a claim of optimality.

\paragraph{Final Output.}
After convergence or reaching $N_{\max}$ rounds, the system outputs the motion from the final iteration:
\[
\mathbf{x}_{\mathrm{final}} = Q(\mathbf{p}_i),
\]
where $i$ denotes the stopping iteration and $Q$ is the MotionLLM executor defined above.
This final motion represents the outcome of a complete self-reflection cycle, incorporating lessons learned from multiple attempts to produce a globally consistent, physically plausible, and semantically grounded interaction sequence.

The overall reflection process is summarized in Algorithm~\ref{alg:memory_reflection}.

\begin{algorithm}[t]
\caption{Memory-Augmented Reflection (with Adaptive Stopping)}
\label{alg:memory_reflection}
\begin{algorithmic}[1]
\Require Scene $\mathbf{s}$; task description $\mathbf{t}$; initial plan $\mathbf{p}_0$;
         MotionLLM executor $Q$; MLLM-based evaluator $f_{\mathrm{MLLM}}$;
         max rounds $N_{\max}$.
\Ensure Trajectory memory $\mathcal{M}_{\mathrm{final}}$ containing all attempts

\State $\mathbf{x}_0 \gets Q(\mathbf{p}_0)$
\State $\mathbf{v}_0 \gets \mathrm{Render}(\mathbf{x}_0)$
\State $\mathcal{M}_0 \gets \{(\mathbf{p}_0, \mathbf{v}_0)\}$

\For{$i = 1$ to $N_{\max}$}
    \State $\mathbf{p}_i \gets f_{\mathrm{reflect}}(\mathcal{M}_{i-1}, \mathbf{s}, \mathbf{t})$
    \State $\mathbf{x}_i \gets Q(\mathbf{p}_i)$
    \State $\mathbf{v}_i \gets \mathrm{Render}(\mathbf{x}_i)$
    \State $\mathcal{M}_i \gets \mathcal{M}_{i-1} \cup \{(\mathbf{p}_i,\mathbf{v}_i)\}$
    \State $\text{status}_i \gets f_{\mathrm{MLLM}}(\mathbf{p}_i, \mathbf{v}_i, \mathbf{t})$
    \If{$\text{status}_i = \text{Pass}$}
        \State \textbf{break} \Comment{Task requirements satisfied}
    \EndIf
\EndFor

\State $\mathcal{M}_{\mathrm{final}} \gets \mathcal{M}_i$
\State \Return $\mathcal{M}_{\mathrm{final}}$
\end{algorithmic}
\end{algorithm}

Together, Online Observation and Memory-Augmented Reflection organize feedback at two temporal scales: local recovery within an attempt and trajectory-level revision across attempts. Memory is used as inference-time context; no model parameters are updated. The contribution lies in how execution evidence is incorporated into the planning and recovery loop, rather than in a new memory storage architecture.

\section{Experiments}

We evaluate our training-free reflection framework on language-driven, long-horizon human--scene interaction tasks.
Based on the structure of the framework and the nature of our tasks, we select representative datasets and metrics to assess both task-level success and physical plausibility.
All experiments are designed to verify how reflection improves the quality and consistency of long-horizon human--scene interactions.

\subsection{Datasets and Metrics}
Previous human--scene interaction datasets mainly focus on short and constrained interactions~\cite{hassan2021stochastic,peng2021amp,zhao2022compositional}, often limited to single-object contact or short motion clips without long-horizon reasoning.
To comprehensively evaluate our training-free reflection framework, we draw from two complementary resources.

TRUMANS~\cite{jiang2024scaling} provides physically detailed indoor scenes with dynamic object configurations, supporting realistic assessment of motion plausibility.

UniHSI~\cite{xiao2023unified} introduces language-driven, arbitrary-length interaction tasks, aligning closely with our text-conditioned planning structure.

While these datasets originally target different tasks, we adapt their scene layouts and object configurations to fit our reflection framework, focusing on the ability to interpret textual instructions and generate coherent long-horizon motions.
All motions are represented in the SMPL-X format and visualized within the reconstructed 3D scenes.

For quantitative evaluation, we assess the framework at both the task level and the motion level.
Following Hassan et al.~\cite{hassan2023synthesizing}, we use Execution Success Rate (ESR) to measure whether each interaction task is successfully completed.
To further evaluate the overall physical quality of generated motion sequences, we adopt Physical Validity (PV),
which measures the extent of body--scene penetration~\cite{zhao2023synthesizing} throughout the sequence.

\begin{equation}
s_{\text{inter-pen}} = \sum_{v_i \in V} -\left(\Psi_O(v_i)\right)_-
\end{equation}

where $\Psi_O$ is the object signed distance field, $(\cdot)_{-}$ clips all positive distance values to zero, and $V$ is the set of body vertices. In our setting, the number of frames varies significantly 
across tasks, and long-horizon sequences often contain large portions 
of non-interactive movement (e.g., walking or navigation before actual 
interaction). Directly averaging $s_{\text{inter-pene}}$ over all frames  
would therefore dilute meaningful penetration events.

To obtain a more reliable measure of physical consistency, we report:
(1) the \textbf{maximum penetration} across all frames, and  
(2) the \textbf{top-$k$ mean} penetration among the highest-penetration 
frames.  
These two statistics capture the severity and the persistence of 
penetration respectively, providing a more robust evaluation for 
long-horizon interaction tasks with highly imbalanced interactive durations.

Together, ESR and PV jointly capture both how well the task is achieved and how physically plausible the resulting motion is.

\subsection{Implementation Details}
Unless otherwise specified, we instantiate the multimodal large
language model (MLLM) with GPT-4o via its official API.
All scene-analysis, online-observation, and reflection prompts share
a unified JSON-based output format. No task-specific finetuning is
performed on either the MLLM or the motion backbone.

For motion synthesis, we adopt the public MotionLLM executor from the
Motion-Agent framework, which uses a Gemma-2B language backbone, a
VQ-VAE motion decoder, and SMPL-X reconstruction. We keep all motion
hyperparameters and codebook settings from the released implementation.
Action descriptions are passed as natural-language sequences and decoded
into SMPL-X parameters at 30 FPS. Rendering is performed in the TRUMANS
and UniHSI scenes.

For reflection, we set the maximum number of rounds
$N_{\max}=5$ unless otherwise noted. For Online Observation, each
atomic action allows up to $R_{\max}=2$ local retries. Experiments
run on a single NVIDIA A6000 GPU, with motion generation time dominated
by rendering and MLLM inference.

\paragraph{Main Result.}

\begin{table}[t]
\centering
\small
\setlength{\tabcolsep}{4pt}
\begin{tabular}{l|ccc|c|c}
\toprule
\multirow{2}{*}{\textbf{Methods}} &
\multicolumn{3}{c|}{\textbf{Success Rate (\%) $\uparrow$}} &
\textbf{pe$_{\text{max}}$ $\downarrow$} &
\textbf{pe$_{10\%}$ $\downarrow$} \\
\cmidrule(lr){2-4}
& Low & Mid & High & & \\
\midrule
UniHSI-CoCo        & \textbf{85.1} & 47.7 & 17.7 & \textbf{18.2} & \textbf{9.3} \\
\midrule
Naive         & 25.5 & 18.2 & 4.4  & 528.4 & 113.6 \\
Online        & 74.4 & 52.3 & 31.1 & 102.7 & 54.5 \\
Ref + Offline & 72.3 & 47.7 & 26.7 & 62.3  & 36.8 \\
\midrule
\textbf{InteractAgent}
& 84.3
& \textbf{79.5}
& \textbf{64.4}
& 48.6
& 29.7 \\
\bottomrule
\end{tabular}
\caption{\textbf{Result of methods on tasks of different complexity levels.}
Success Rate (SR, \%) and penetration errors are reported.}
\label{tab:baseline_ablation}
\end{table}

Table~\ref{tab:baseline_ablation} compares InteractAgent with the supervised UniHSI-CoCo baseline and training-free component variants. Online Observation increases success rates from 25.5/18.2/4.4\% to 74.4/52.3/31.1\% on low-/medium-/high-complexity tasks, respectively. This comparison supports the benefit of step-wise feedback for reducing execution errors.

\paragraph{Individual effects of core modules.}
The Ref + Offline variant reduces $\mathrm{pe}_{\max}$ from 528.4 to 62.3, approximately an 8.5-fold reduction relative to Naive. Combining reflection with Online Observation further improves success rates to 84.3/79.5/64.4\% and reduces $\mathrm{pe}_{\max}$ to 48.6. These results indicate complementary benefits from local feedback and cross-attempt refinement.

\paragraph{Full model performance and analysis.}
Among the training-free variants in Table~\ref{tab:baseline_ablation}, InteractAgent achieves the highest success rate at each complexity level and the lowest penetration scores. Compared with UniHSI-CoCo, InteractAgent improves medium- and high-complexity success rates by 31.8 and 46.7 percentage points, respectively. UniHSI-CoCo retains a slightly higher success rate on low-complexity tasks and lower penetration scores. The results therefore support improved long-horizon task completion, while also showing that visual feedback does not replace the geometric precision of the supervised baseline.

\paragraph{Ablation Study.}

\paragraph{Ablation on LLM Planner.}
We study how the choice of multimodal LLM affects planning robustness.
Table~\ref{tab:llm_ablation} compares GPT-4o and Qwen-VL-Max under identical settings.
GPT-4o yields higher Planning Correctness (PC) and SR, while Qwen-VL-Max provides competitive grounding but struggles more on complex multi-step reasoning.

\begin{table}[t]
\centering
\begin{tabular}{lcc}
\toprule
LLM Type & SR (\%) $\uparrow$ & PC (\%) $\uparrow$ \\
\midrule
Human     & 72 & --   \\
w. Qwen-VL-Max p & 56 & 75 \\
w.  GPT-4o   & 64 & 81 \\
\bottomrule
\end{tabular}
\caption{Comparison of SR and PC (Y/N decided by human) across different LLM types.}
\label{tab:llm_ablation}
\end{table}

\paragraph{Ablation on MotionLLM Executor.}
We compare the default MotionLLM with alternative executor configurations while retaining the high-level InteractAgent framework (Table~\ref{tab:motion_executor_ablation}). MG-MotionLLM~\cite{wu2025mg} achieves a success rate of 72\%, compared with 70\% for the default MotionLLM, and reduces both reported penetration scores. The MG stateless configuration achieves 70\% success with higher penetration than MG-MotionLLM. These results show that the motion executor affects the quality of the generated trajectories; they do not by themselves isolate a single bottleneck in task completion.

\begin{table}[t]
\centering
\begin{tabular*}{\columnwidth}{@{\extracolsep{\fill}}lccc}
\toprule
Motion Executor & SR (\%)\textuparrow & pe$_{\text{max}}$\textdownarrow & pe$_{10\%}$\textdownarrow \\
\midrule
MotionLLM     & 70 & 48.6 & 29.7 \\
MG-MotionLLM  & 72 & 23.3 & 16.9 \\
MG stateless  & 70 & 46.2 & 24.3 \\
\bottomrule
\end{tabular*}
\caption{Comparison of SR and penetration between Motion Executors.}
\label{tab:motion_executor_ablation}
\end{table}

\paragraph{Summary.}
The component comparisons support complementary roles for Online Observation and Memory-Augmented Reflection. The backbone comparisons additionally show that performance depends on the pretrained planner and motion executor. Within the training-free variants reported in Table~\ref{tab:baseline_ablation}, the full framework provides the strongest task success and lowest penetration scores.

\section{Limitations and Future Work}
InteractAgent depends on the capabilities of both the pretrained MLLM and the frozen motion executor. Planning errors, inaccurate visual judgments, and limited action-to-motion fidelity can remain after refinement. Human-readable plans and feedback facilitate inspection, but do not establish faithful interpretability of the underlying models.

Visual evaluation is also limited by rendering viewpoints, occlusion, and the temporal sampling of motion. It may miss collisions or contact errors that are detectable using 3D geometry. Accordingly, task success and geometric penetration should be considered complementary measures. Our results do not establish that the generated motions satisfy all physical constraints.

Reflection introduces additional model calls, motion generation, and rendering. The retry and reflection limits bound the number of correction attempts, but the selected budgets are not established as optimal. Moreover, storing prior planning--execution pairs increases the context supplied to later reflection calls. Future work may investigate budget allocation, memory compression, viewpoint robustness, and stronger motion executors.

\section{Conclusion}
We presented InteractAgent, a training-free framework for long-horizon human--scene interaction that couples a pretrained MLLM with a frozen MotionLLM executor. Online Observation provides step-wise recovery, while Memory-Augmented Reflection uses previous planning--execution pairs to revise subsequent attempts. The reported experiments support the complementary benefits of these mechanisms for task completion and penetration reduction relative to the evaluated training-free variants. InteractAgent exposes human-readable intermediate plans and feedback, while remaining subject to the limitations of visual evaluation and pretrained motion generation.

\clearpage

{
    \small
    \bibliographystyle{ieeenat_fullname}
    \bibliography{main}
}

\clearpage
\onecolumn
\appendix

\section{Datasets and Metrics}
\label{app:metrics}

\subsection{Datasets}
Previous human--scene interaction datasets mainly focus on short and constrained interactions, often limited to single-object contact or short motion clips without long-horizon reasoning.
To comprehensively evaluate our training-free reflection framework, we draw from two complementary resources.

TRUMANS provides physically detailed indoor scenes with dynamic object configurations, supporting realistic assessment of motion plausibility.

UniHSI introduces language-driven, arbitrary-length interaction tasks, aligning closely with our text-conditioned planning structure.

While these datasets originally target different tasks, we adapt their scene layouts and object configurations to fit our reflection framework, focusing on the ability to interpret textual instructions and generate coherent long-horizon motions.
All motions are represented in the SMPL-X format and visualized within the reconstructed 3D scenes.

\subsection{Metrics}

\subsubsection{ESR}
For quantitative evaluation, we assess the framework at both the task level and the motion level.
Following Hassan et al., we use Execution Success Rate (ESR) to measure whether each interaction task is successfully completed.

\subsubsection{PV}
To further evaluate the overall physical quality of generated motion sequences, we adopt Physical Validity (PV), which measures the extent of body--scene penetration throughout the sequence following prior work~\cite{zhao2023synthesizing}. Let $\Psi_O$ denote the object signed distance field (SDF), $(\cdot)_{-}$ retain negative distance values while clipping positive values to zero, and $V$ denote the set of body vertices. The penetration score for each frame is defined as
\begin{equation}
s_{\text{inter-pene}}
= -\sum_{v_i \in V}\left(\Psi_O(v_i)\right)_{-}
\end{equation}
The negative sign converts the accumulated negative SDF values into a non-negative penetration score, where a larger value indicates more severe body--scene penetration.

In our setting, sequence lengths vary substantially across tasks, and long-horizon motions often contain large portions of non-interactive movement, such as walking or navigation before the actual interaction. Directly averaging $s_{\text{inter-pene}}$ over all frames would therefore dilute meaningful penetration events and make results sensitive to sequence length.

To obtain a more reliable measure of physical consistency, we report two complementary statistics: (1) the \textbf{maximum penetration}, pe$_{\text{max}}$, across all frames, and (2) the \textbf{top-$10\%$ mean penetration}, pe$_{10\%}$, computed over the frames with the highest penetration scores. The former captures the severity of the worst physical violation, while the latter measures whether substantial penetration persists across multiple frames rather than occurring as an isolated event. These statistics provide a more robust evaluation for long-horizon interaction tasks with highly imbalanced interactive and non-interactive durations.

Together, Execution Success Rate (ESR) and Physical Validity (PV) evaluate both whether the instructed task is successfully completed and whether the generated motion remains physically plausible within the scene.

\subsubsection{Cost metrics}

Since InteractAgent is a modular framework whose actual latency depends on the selected MLLM, motion executor, hardware, and API conditions, we report hardware-independent operation counts as proxies for computational cost. We decompose the framework overhead into three components: multimodal queries, motion-model executions, and trajectory rendering.

We define one \textbf{visual-query unit (VQU)} as one MLLM request containing a prompt and one rendered trajectory observation. This normalization accounts for the amount of visual evidence processed rather than the implementation-specific number of API requests. For example, evaluating \(K\) candidate trajectories in Greedy or Beam Search is counted as \(K\) VQUs, even if the candidates can technically be batched into a single API request, because such a request processes substantially more visual content than a standard single-observation query.

A \textbf{motion execution} denotes one invocation of MotionLLM, including motion-token generation and VQ-VAE decoding. This count provides an executor-independent proxy because absolute runtime varies with the selected motion model and hardware. 

A \textbf{rendering operation} denotes the generation of one trajectory visualization. In our implementation, rendering primarily involves scene loading and initialization, while the marginal cost of adding sampled keyframes is relatively small. Using simplified human meshes without complex materials, rendering trajectories containing 10–30 keyframes differs in runtime by only approximately \(1.2\!\times\)–\(1.4\!\times\).

Because each visual query in the evaluated methods is paired with a newly rendered trajectory, the numbers of rendering operations and VQUs are nearly identical. We therefore report VQUs and motion executions as the primary framework-level cost proxies, while noting that VQUs also closely reflect the associated rendering overhead.

\section{Qualitative Examples}
\label{app:qualitative}

\subsection{Online Observation Example}

\begin{center}
\includegraphics[width=\linewidth]{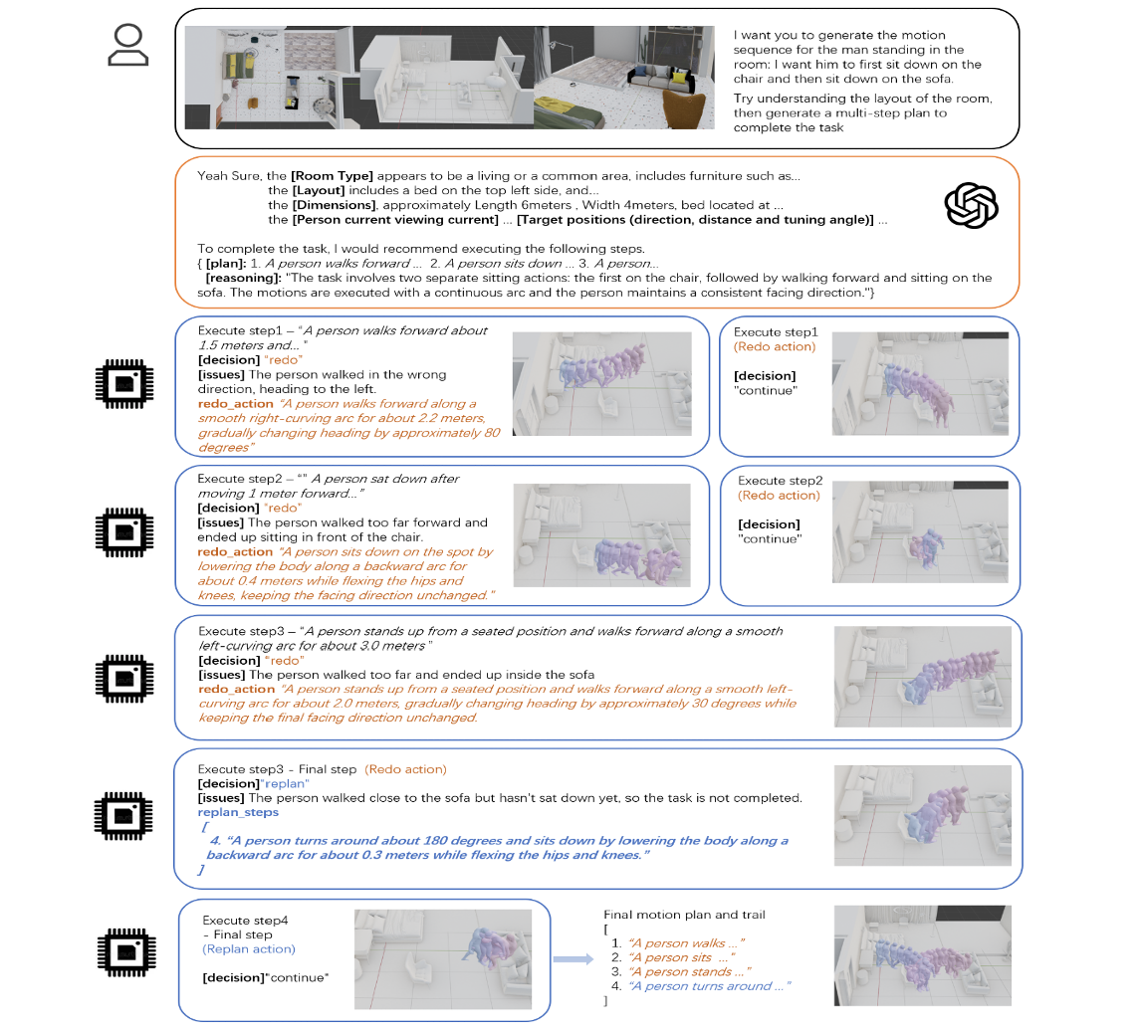}

\vspace{0.3em}
\textbf{Figure B.1: Online Observation example.}
\end{center}

\clearpage

\begin{center}
\includegraphics[width=\linewidth]{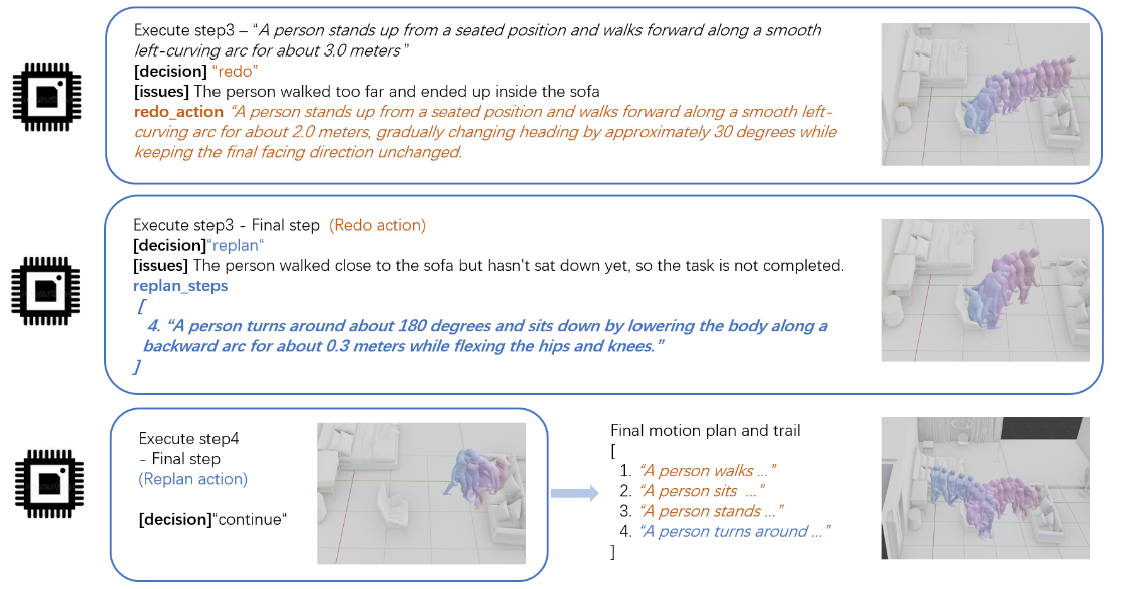}

\vspace{0.3em}
\textbf{Figure B.2: Online Observation example (steps 3--4 and final plan).}
\end{center}

\section{Prompt Templates}
\label{sec:appendix_prompts}

\begin{Verbatim}
Please analyze these two images: the first is a bird's-eye view,
the second is a first-person view.

Please provide the following information:

1. Room type and layout
2. Room dimensions (length x width, in meters)
3. Current viewing direction (facing where)
4. Distance from current position to various furniture (in meters)
5. Required turning angle to reach target position

Important: The first-person view shows your current facing direction,
which is the starting point for motion planning. Please accurately
determine the current orientation and calculate the required turning
angle to reach the target position.

Note: In first-person view, what you see ahead is your forward direction
(0 degrees), left is -90 degrees, right is +90 degrees. Please calculate
turning angles based on this coordinate system.

Please analyze this scene image and provide a detailed description including:

1. Spatial layout: Room type and overall layout
2. Key objects: Key objects and furniture in the scene
3. Dimensions: Room dimensions and object sizes (in meters)
4. Walkable areas: Walkable areas and paths
5. Obstacles: Obstacles and their positions
6. Height info: Height information for objects and surfaces
7. Starting position: Suggest a reasonable starting position for the person
   (e.g., "entrance", "center of room", "near the door")
8. Spatial metrics for planning:
   - Approximate distances between key objects (in meters)
   - Clearance widths of narrow passages (in meters)
   - Relative angular directions to target objects (degrees)
   - Whether movement requires sidestepping, turning in place, or wall-following
9. Navigational affordances:
   - Which paths are straight-walkable
   - Which paths require body rotation or lateral movement
   - Which areas are not passable by normal walking

Please provide a comprehensive analysis that can be used for motion planning
and trajectory generation.
\end{Verbatim}

\end{document}